\documentclass[letterpaper]{article} 
\usepackage{aaai24}  
\usepackage{times}  
\usepackage{helvet}  
\usepackage{courier}  
\usepackage[hyphens]{url}  
\usepackage{graphicx} 
\usepackage{natbib}  
\usepackage{caption} 
\usepackage{algorithm}
\usepackage{algorithmic}

\usepackage{newfloat}
\usepackage{listings}
\DeclareCaptionStyle{ruled}{labelfont=normalfont,labelsep=colon,strut=off} 
\floatstyle{ruled}
\newfloat{listing}{tb}{lst}{}
\floatname{listing}{Listing}
\usepackage{booktabs}
\usepackage{multirow}
\usepackage{amsmath}
\usepackage{amssymb}
\usepackage{color}
\usepackage{makecell}
\usepackage{subcaption}

\usepackage[capitalize]{cleveref}
\crefname{section}{Sec.}{Secs.}
\Crefname{section}{Section}{Sections}
\Crefname{table}{Table}{Tables}
\crefname{table}{Tab.}{Tabs.}

\newcommand{\eqnref}[1]{Eq.~\eqref{#1}}
\newcommand{\secref}[1]{Sec.~\ref{#1}}

\newcommand{\proposed}{UVAGaze}
\newcommand{\upscore}[1]{\scriptsize{\mydarkgreen{$\blacktriangledown$ #1\%}}}
\newcommand{\downscore}[1]{\scriptsize{\mydarkred{$\blacktriangle$ #1\%}}}
\newcommand{\mydarkred}[1]{\textcolor[rgb]{0.8,0.0,0.0}{ #1}}
\newcommand{\mydarkgreen}[1]{\textcolor[rgb]{0.17,0.56,0.36}{ #1}}
\newcommand{\mono}{Mono}
\newcommand{\so}{Dual-S}
\newcommand{\avg}{Dual-A}
\newcommand{\hp}{HPose}

\newcommand{\ie}{\textit{i.e.}}
\newcommand{\eg}{\textit{e.g.}}

\title{UVAGaze: Unsupervised 1-to-2 Views Adaptation for Gaze Estimation}
\author {
    Ruicong Liu,
    Feng Lu\thanks{Corresponding Author}
}
\affiliations {
    State Key Laboratory of VR Technology and Systems, School of CSE, Beihang University, Beijing, China\\
    \{liuruicong, lufeng\}@buaa.edu.cn
}

\usepackage{bibentry}

\begin{document}
\nocopyright
\maketitle

\begin{abstract}
Gaze estimation has become a subject of growing interest in recent research. Most of the current methods rely on single-view facial images as input. Yet, it is hard for these approaches to handle large head angles, leading to potential inaccuracies in the estimation. To address this issue, adding a second-view camera can help better capture eye appearance. However, existing multi-view methods have two limitations.  1) They require multi-view annotations for training, which are expensive.  2) More importantly, during testing, the exact positions of the multiple cameras must be known and match those used in training, which limits the application scenario. To address these challenges, we propose a novel 1-view-to-2-views (1-to-2 views) adaptation solution in this paper, the Unsupervised 1-to-2 Views Adaptation framework for Gaze estimation (\proposed). Our method adapts a traditional single-view gaze estimator for flexibly placed dual cameras. Here, the ``flexibly" means we place the dual cameras in arbitrary places regardless of the training data, without knowing their extrinsic parameters. Specifically, the \proposed~builds a dual-view mutual supervision adaptation strategy, which takes advantage of the intrinsic consistency of gaze directions between both views. In this way, our method can not only benefit from common single-view pre-training, but also achieve more advanced dual-view gaze estimation. The experimental results show that a single-view estimator, when adapted for dual views, can achieve much higher accuracy, especially in cross-dataset settings, with a substantial improvement of 47.0\%. Project page: \url{https://github.com/MickeyLLG/UVAGaze}.

\end{abstract}

\section{Introduction}

Gaze estimation is a crucial indicator of user attention and has been widely adopted in various applications, including human-robot interaction \cite{A:admoni2017social,A:terziouglu2020designing,A:wang2015hybrid}, semi-autonomous driving \cite{A:demiris2007prediction,A:majaranta2014eye,A:park2013predicting}, medical diagnostics \cite{A:castner2020deep}, and augmented/virtual reality games \cite{A:burova2020utilizing,A:konrad2020gaze,A:wang2020comparing}. With the advancements in deep learning, numerous gaze estimation networks \cite{G:krafka2016eye,G:park2018deep,G:zhang2022gazeonce,G:zhang2017s,A:wang2023wheelchair} and adaptation/optimization methods \cite{D:bao2022generalizing,G:cheng2022puregaze,D:liu2021generalizing,G:park2019few,D:yu2019improving} have been proposed in the last decade.

However, the majority of existing gaze estimation methods still use single-view facial images as input. As shown in \cref{tab:diff}, although using single-view facial images is simple and effective for learning-based approaches, it poses several challenges. A drawback of traditional single-view gaze estimation is its limited reliable output range for head poses. Empirically and experimentally, 
when the camera is facing the side of the face, the testing accuracy is significantly reduced due to occlusion and deformation problems.
\begin{figure}
	\centering
	\includegraphics[width=\linewidth]{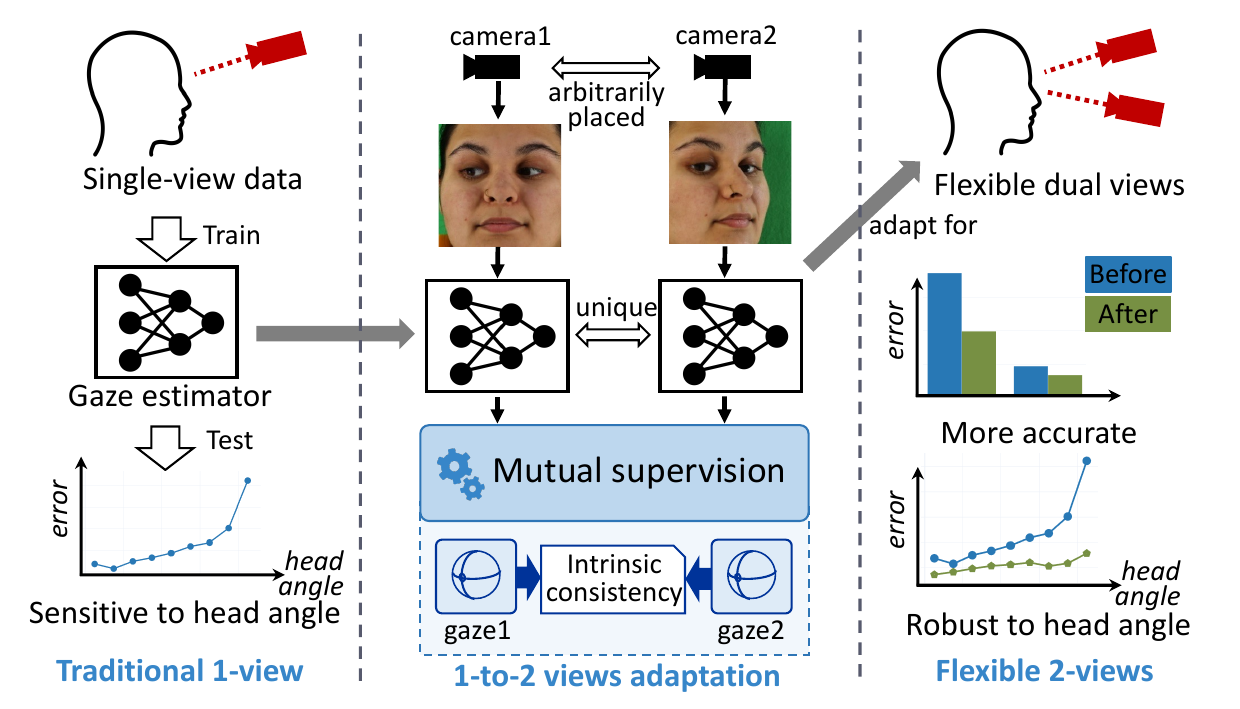}
	\caption{Overview of the proposed unsupervised 1-view-to-2-views adaptation method for gaze estimation. A single-view estimator can be adapted for flexible dual views.} 
	\label{fig:teaser}
\end{figure}

\begin{table*}[t]
\small
\begin{center}
\renewcommand\arraystretch{.85}
\setlength{\tabcolsep}{.5mm}
{\small
\begin{tabular}{l|ccc}
\bottomrule[1.2pt]
\specialrule{0em}{1pt}{1pt}
Method & Training & Test  & Output\\
\hline
\specialrule{0em}{1pt}{1pt}
\makecell[l]{Traditional} & \makecell[c]{\textbf{1}-view   (common data)} & \makecell[c]{\textbf{1}-view   (arbitrary camera pose)} & \makecell[c]{1-view   (\mydarkred{limited} head pose range)}\\
\hline
Two-views training&  \makecell[c]{\textbf{2}-views  (\mydarkred{multi-view} data)}  & \makecell[c]{\textbf{2}-views  (\mydarkred{same} camera poses with training)} & Select/Average/etc. (\mydarkgreen{extended} head pose range)  \\
\hline
Ours (Adaptation) &  \makecell[c]{\textbf{1}-view (\mydarkgreen{common} data)} & \makecell[c]{\textbf{2}-views (\mydarkgreen{arbitrary} camera poses)} &  \makecell[c]{Select/Average/etc.  (\mydarkgreen{extended} head pose range)} \\
\bottomrule[1.2pt]
\end{tabular}  }
\end{center}
\caption{Comparison of our method with traditional single-view gaze estimation and existing multi-view methods 
that require at least two-views training (\eg, feature concatenation).
``Select/Average/etc." indicates the strategy to produce the gaze output. In particular, ``Select" means selecting one view for the output, ``Average" means averaging the predictions from the two views.
}
\label{tab:diff}
\end{table*}

To address this challenge, a natural solution is to add a second-view camera to better capture the eye appearance. In this paper, we also demonstrate the benefits of using dual cameras. Concurrently, several existing studies have paid attention to gaze estimation in multi-view settings \cite{M:arar2017robust,M:kim2020preliminary,M:lian2018multiview,M:cheng2023dvgaze}. However, all these methods require training with at least two views. Their models consistently utilize input images from multiple cameras during both training and testing, predominantly employing feature concatenation or fusion to leverage multi-view information. Such methodologies exhibit two main limitations, as outlined in \cref{tab:diff}: 
1) They require costly multi-view annotations for training, especially for the feature concatenation or fusion module. 
2) Crucially, during testing, the positions and poses of the multiple cameras must be known and match those employed in training, thus limiting the application scenarios.

Unlike those existing methods, this paper proposes a novel unsupervised 1-view-to-2-views adaptation solution. Our approach requires only a pre-trained estimator and unlabeled input images from flexible dual views.
By ``flexible", we mean that the dual cameras can be placed arbitrarily, regardless of the pre-training data, without the needing to know their specific extrinsic parameters.
As shown in \cref{tab:diff}, our method can not only benefit from common single-view pre-training, but also adapt a single-view estimator for more advanced dual-view inference, resulting in reliable outputs in an extended head pose range.

In this paper, an Unsupervised 1-to-2 Views Adaptation framework for Gaze estimation (\proposed) is proposed.
To adapt a single-view estimator for dual-view inference, we leverage the intrinsic consistency of the gaze directions between the two views. Specifically, our method builds a mutual supervision strategy that dynamically judges the reliability of each view, allowing the predictions from one view to supervise the other. 
The primary contributions of this paper are summarized as follows.
\begin{itemize}
	\setlength{\itemsep}{0pt}
	\setlength{\parsep}{0pt}
	\setlength{\parskip}{0pt}
	\item For the first time, we propose a novel unsupervised 1-view-to-2-views adaptation (1-to-2 views UVA) framework for gaze estimation. A traditional single-view estimator can be adapted for flexible dual views without gaze annotations or camera extrinsic parameters.
	\item We design a dual-view mutual supervision strategy for the 1-to-2 views UVA. It utilizes the intrinsic consistency of gaze directions and can dynamically judge the reliability of each view.
	\item Empirical evaluation highlights the advantage of \proposed~in adapting a single-view estimator, particularly in cross-dataset scenarios, with a notable 47.0\% improvement over traditional single-view methods.
\end{itemize}

\section{Related Work}
\subsection{Appearance-Based Gaze Estimation}
Appearance-based gaze estimation predicts 3D gaze directions from facial images. Various networks have been designed for better accuracy, such as adversarial-learning \cite{D:wang2019gazeadv}, two-eye-asymmetry \cite{G:cheng2018appearance}, and coarse-to-fine \cite{G:cheng2020coarse} structures. Recently, studies \cite{D:bao2022generalizing,G:cheng2022puregaze,D:liu2021generalizing,G:kellnhofer2019gaze360} have utilized ResNet \cite{N:he2016deep}, inspired by previous work \cite{G:zhang2020eth} that proves its effectiveness for gaze estimation.

To achieve better accuracy, increasing research is focusing on aspects like cross-dataset performance \cite{D:liu2021generalizing,D:bao2022generalizing} and generalization \cite{G:cheng2022puregaze}. 
However, these studies predominantly approach gaze estimation as a single-view task, ignoring the potential challenges posed by large head angles. 
In this paper, we will design a method to extend gaze estimation to dual views.

\subsection{Multi-View Gaze Estimation} \label{sec:multi-view}
Multi-view gaze estimation remains a relatively unexplored research area. Prior work \cite{M:gideon2022unsupervised} learns effective gaze representations from multiple views but can be only used for single-view inference. Some studies \cite{M:arar2017robust,M:kim2020preliminary,M:lian2018multiview} aim to train more accurate gaze estimators using data from multiple views.
However, the aforementioned studies share two drawbacks. 1) They require multi-view annotations to train the models. It is expensive and challenging to collect reliable gaze data under multiple views. 2) During testing, the poses of the multiple cameras are assumed to be known and covered by the training data, hereby limiting their applicability. In contrast, our unsupervised view adaptation (UVA) method eliminates the need for multi-view annotations and is adaptable to arbitrarily placed dual views.

\begin{figure}
\centering
\includegraphics[width=.9\linewidth]{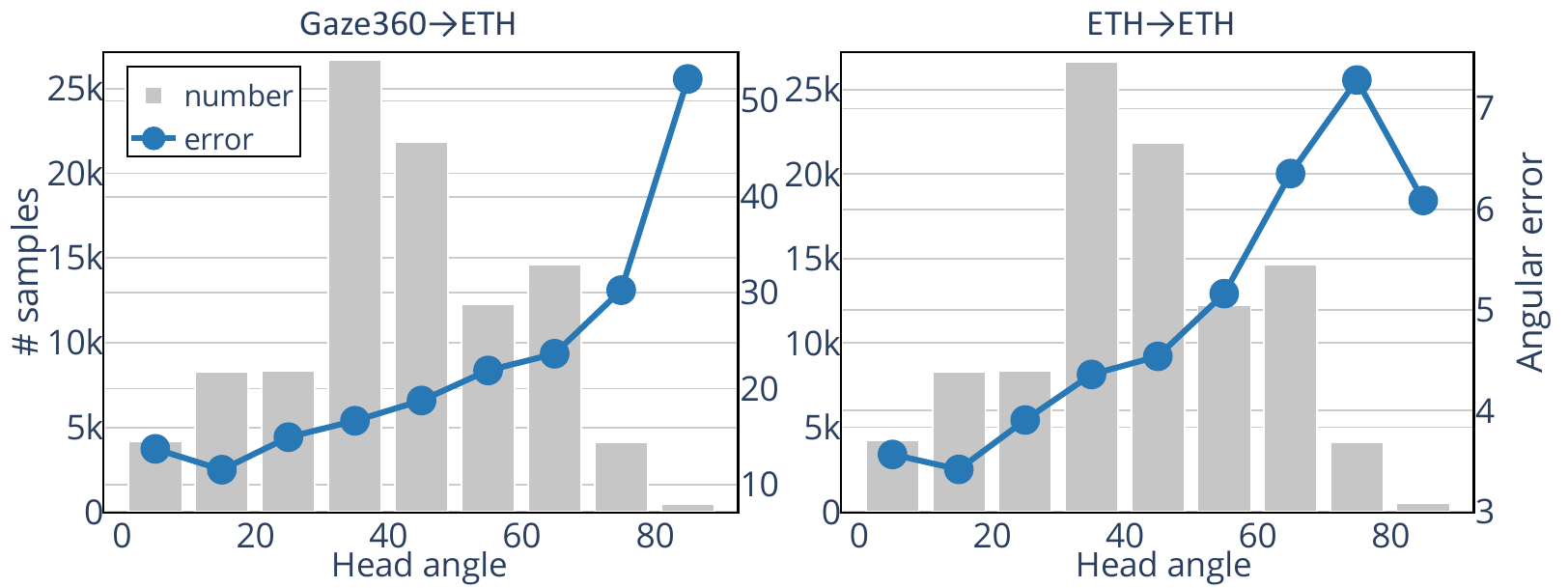}
\caption{Number of samples and gaze errors with respect to head angles. Left: the model is trained on Gaze360 \cite{G:kellnhofer2019gaze360} dataset and tested on ETH-XGaze \cite{G:zhang2020eth}. Right: both the training and testing datasets are ETH-XGaze.}
\label{fig:tendency}
\end{figure}


\section{Motivation}

\subsection{Gaze Estimation Using Two Views}
\label{sec:observation}
Upon analyzing the error of traditional single-view gaze estimation, we have observed the error increases as the head angle increases (capturing the side of the face), as illustrated in \cref{fig:tendency}. This phenomenon can be attributed to the fact that large head angles can cause occlusion and deformation.  
Therefore, introducing a second-view camera could enhance the capture of eye appearance. 

As stated in \cref{tab:diff}, existing multi-view methods are not practical, since they have limitations in the training data and camera poses. Therefore, we plan to propose an novel unsupervised view adaptation (UVA) method, which is the first to adapt a gaze estimator from 1-view to 2-views.

\subsection{Output of Two-Views Based Gaze Estimation}
\label{sec:intuitive}
As shown in \cref{tab:diff}, to calculate the gaze estimation results based on dual views, there are two typical approaches: select-front and average. Given predictions from both views, we can either 1) select the prediction with a smaller head angle as the input for estimation or 2) transform the gaze predictions from both views to the same coordinate system and average them. In \cref{tab:validation}, we evaluate both approaches by applying them to a single-view baseline estimator. 
Specifically, we use the dual-view data ETH-XGaze dataset \cite{G:zhang2020eth} for testing, and the head pose labels are used for selecting or averaging.

We conduct experiments under two settings: Gaze360 and ETH$\rightarrow$ETH, with the only difference being the pre-training dataset. As indicated in \cref{tab:validation}, both methods enhance accuracy, confirming the benefit of dual-view information. However, this improvement stems only from merging at the final prediction stage. If we fully harness dual-view data for model optimization and combine it with the two methods, we anticipate even higher accuracy.
\begin{table}
\small
\begin{center}
\small
\renewcommand\arraystretch{0.8}
\setlength{\tabcolsep}{1mm}
\begin{tabular}{lcc}  
    \toprule[1.2pt]
    & Gaze360 $\rightarrow$ ETH & ETH $\rightarrow$ ETH  \\ 
    \midrule[1pt]
    Baseline & 18.73 & 4.75 \\ 
    \midrule[1pt]
    Select-front & 16.03 \upscore{14.4} & 4.08 \upscore{14.1} \\
    Average & 17.69 \upscore{5.6} & 4.05 \upscore{14.7} \\
    \bottomrule[1.2pt]
\end{tabular}
\end{center}
\caption{
The gaze estimation results of the two approaches to produce the output based on two views. Angular error is used as the metric.
}
\label{tab:validation}
\end{table}

\subsection{One-to-two Views UVA for Gaze Estimation}
Based on the discussed ideas, we propose an unsupervised 1-view-to-2-views adaptation task (1-to-2 views UVA). 
The aim is to adapt a single-view gaze estimator for dual-view inference in an unsupervised manner.
Compared with existing multi-view methods, this approach offers two main benefits:
1) It avoids the need for multi-view annotations, reducing data collection costs.
2) Dual cameras can be arbitrarily placed regardless of training data, without knowing their extrinsic parameters, which enhances its practicality.
In this way, our method not only benefits from common single-view training, but also achieves advanced dual-view gaze estimation.
To achieve this, we design a mutual supervision strategy that uses the dual-view information to optimize the model,
and the details are explained in \cref{sec:method}.



\begin{figure*}
	\centering
	\includegraphics[width=.88\linewidth]{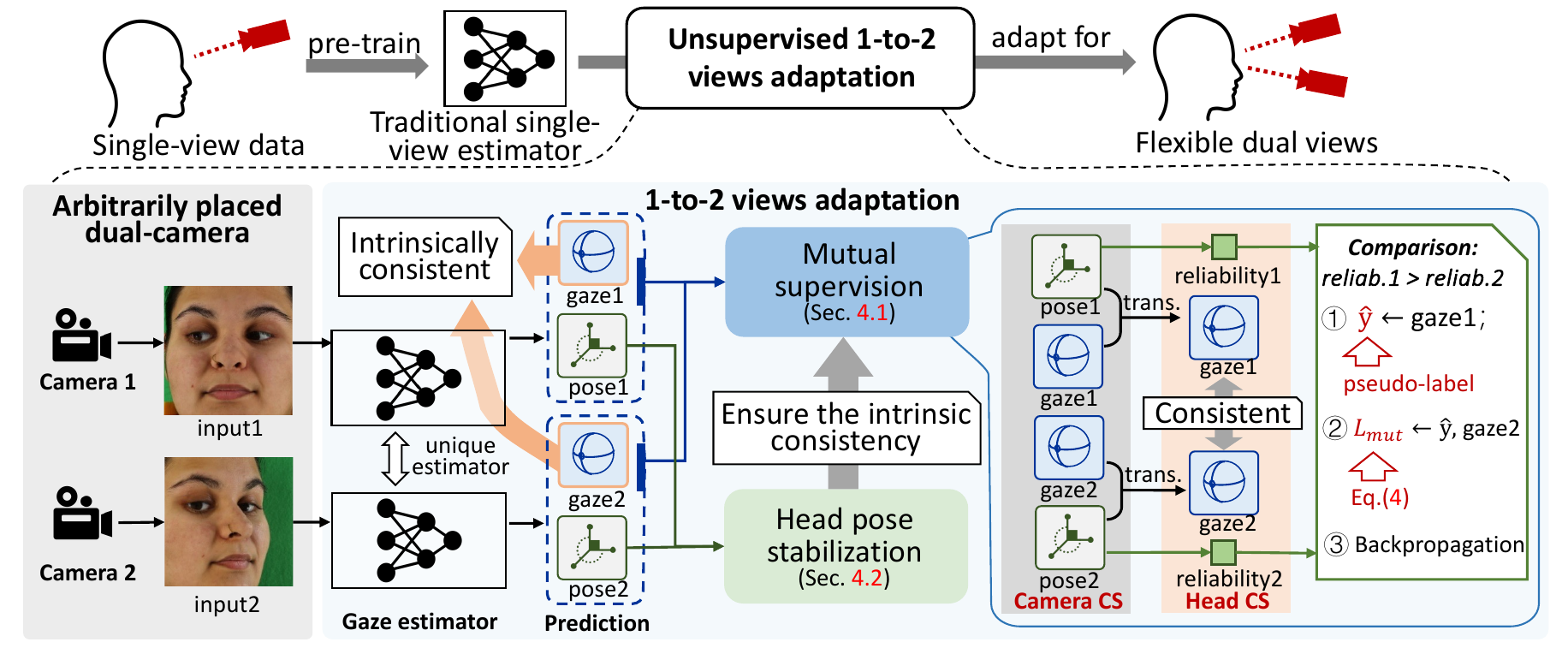}
	\caption{Overview of the proposed \proposed, images captured from an arbitrarily placed dual-camera pair are used for UVA. Their gaze and head pose predictions are then used for mutual supervision. In the mutual supervision, gaze predictions are transformed from the camera coordinate system (Camera CS) into the head coordinate system (Head CS) using head pose predictions. Subsequently, with the reliability generated from the head pose predictions, the loss for mutual supervision is calculated. The core idea is to leverage the intrinsic consistency of gaze directions. And to ensure the consistency, a head pose stabilization module is proposed.}
	\label{fig:overview}
\end{figure*}

\section{UVAGaze: 1-to-2 Views Gaze Estimation}\label{sec:method}
We propose the Unsupervised 1-to-2 Views Adaptation framework for Gaze estimation (\proposed), which adapts a single-view estimator for flexibly placed dual cameras. An overview is shown in \cref{fig:overview}. The central idea is to exploit the intrinsic consistency of the gaze directions under both views. 
To ensure the consistency, we also design a head pose stabilization module, supervised by the fixed rotation transformation between the two cameras. 

\subsection{Mutual Supervision Strategy}\label{sec:mutual}
\textbf{Forward propagation.} The \proposed~ merges the models under both views to be a unique $G(\cdot|\Theta)$, which receives input from the dual-camera. 
The parameters of the model $\Theta$ are pre-trained before adaptation and can predict gaze direction and head pose from an input image $\mathbf{x}$. The forward function is:
\begin{equation} \label{eq:forward}
	\mathbf{g}, (\alpha, \beta, \gamma)=G(\mathbf{x}| \Theta),
\end{equation}
\noindent where $\mathbf{g}$ represents the 3D gaze direction, and $(\alpha, \beta, \gamma)$ denote the yaw, pitch, and roll angles of the head pose, respectively. Note that both gaze and head pose are defined in the camera coordinate system. Building on this, the head angle $\theta$ is defined as the angle between the z-axes of both the head and camera coordinate systems, \ie:
\begin{equation} \label{eq:theta}
	\theta=\langle R(\alpha, \beta, \gamma) \times [0, 0, 1]^T, [0, 0, 1]^T \rangle,
\end{equation}
\noindent where $R$ indicates the rotation matrix, and $\langle \mathbf{a}, \mathbf{b} \rangle$ indicates the angle between $\mathbf{a}$ and $\mathbf{b}$.

\textbf{Mutual supervision.} Since gaze estimation under dual-view is a stereo problem, the two gaze directions should be identical in the head coordinate system, \ie, theoretically $R_1^T\mathbf{g}_1=R_2^T\mathbf{g}_2$ (where the subscript indicates the input camera ID). However, directly imposing $R_1^T\mathbf{g}_1=R_2^T\mathbf{g}_2$ during adaptation may not be optimal. As discussed in \cref{sec:observation}, the reliability of predictions varies with head angles. Therefore, our method dynamically judges the reliability of gaze predictions based on their head pose predictions. The more reliable gaze predictions serve as pseudo-labels to supervise consistency.
This process is defined as follows:
\begin{equation} \label{eq:p-y}
	(p, \mathbf{\hat{y}})=\left\{
	\begin{aligned}
		(0, R_1^T\mathbf{g}_1) & , & \theta_1 \leq \theta_2, \\
		(1, R_2^T\mathbf{g}_2) & , & else,
	\end{aligned}
	\right.
\end{equation}
where $\mathbf{\hat{y}}$ is the pseudo-label, and $p$ serves as a flag indicating the more reliable view. For example, $p=0$ means the view from camera 1 can generate more reliable predictions. Note that the pseudo-label is transformed into head coordinate system by left multiplying with $R^T$.

Then, the less reliable gaze predictions are supervised by the pseudo-label. The mutual supervision loss function is:
\begin{equation} \label{eq:mut}
	\mathcal{L}_{mut}=p\langle R_1^T\mathbf{g}_1, stop(\mathbf{\hat{y}}) \rangle + (1-p)\langle R_2^T\mathbf{g}_2, stop(\mathbf{\hat{y}}) \rangle,
\end{equation}
where $stop(\cdot)$ cuts off the back propagation of gradient. 
In this manner, \proposed~dynamically judges the reliability of both views and enables mutual supervision by leveraging the consistency of gaze directions between both views.

\begin{figure}
\centering
\includegraphics[width=.9\linewidth]{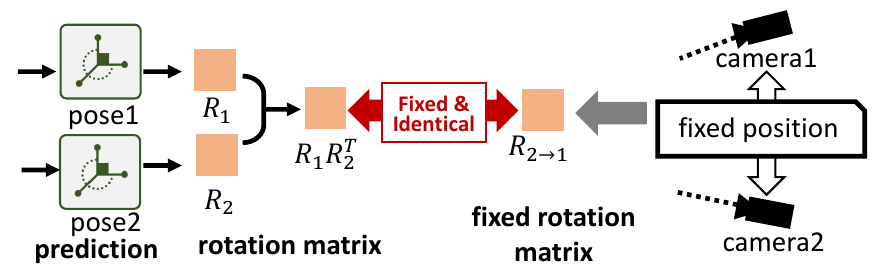}
\caption{The idea of the head pose stabilization module.}
\label{fig:stabilization}
\end{figure}

\subsection{Momentum Head Pose Stabilization}\label{sec:stabilization}
Unfortunately, during mutual supervision, the consistency of gaze may fail, and error amplification can occur easily since the model trains itself by using its own predictions. Previous work \cite{D:liu2021generalizing} has also described this phenomenon. To ensure consistency, we design a module to stabilize head pose predictions. This is because it is the head pose predictions that are used by mutual supervision to transform the gaze predictions to the head coordinate system. 

In this module, as shown in \cref{fig:stabilization}, we use the fixed positional relationship between the dual-camera to stabilize the head pose predictions. Since the positions of the two cameras are fixed, the rotation matrix between the two camera coordinate systems is also constant, \ie, $R_1R_2^T=constant$. From this formula, we can draw the following equation:
\begin{equation} \label{eq:constant}
	\begin{aligned}
		C = & f(\alpha_1, \alpha_2, \beta_1, \beta_2)= \sin\beta_1\sin\beta2\\
		& +\sin\alpha_1\sin\alpha_2\cos\beta_1\cos\beta_2 \\
		& + \cos\alpha_1\cos\alpha_2\cos\beta_1\cos\beta_2.
	\end{aligned}
\end{equation}
\noindent \eqnref{eq:constant} computes the value of the (3, 3) element of the
rotation matrix $R_1R_2^T$. We use it to represent the entire matrix and check its stability.
Note that the roll angle $\gamma$ is not introduced, because the effect of $\gamma$ is eliminated in the normalization step of the input image. Also, its predicted values are often in the [-0.01, 0.01] interval, having no effect on the task. The normalization step is described in \cref{sec:preparation}.

In practice, the input facial image is normalized using an affine transformation matrix $W$, derived from the head pose and camera focal length. Thus, our method initially performs an inverse transformation on $R_1$ and $R_2$, leading to a revised equation: $(W_1^{-1}R_1)(W_2^{-1}R_2)^T=C$.

To stabilize the head pose predictions using the constant $C$, we construct a momentum variable during adaptation by temporal averaging. Specifically, for the current iteration $T$, the variable is denoted as $C^{(T)}$ and can be updated as:
\begin{equation} \label{eq:momentum}
	C^{(T+1)} = \eta C^{(T)} + (1-\eta) f^{(T)}(\alpha_1, \alpha_2, \beta_1, \beta_2),
\end{equation}
\noindent where $C^{(T+1)}$ represents the momentum variable for the next iteration. $\eta$ represents the momentum, which is usually set to 0.99.
The momentum variable is employed to stabilize the head pose predictions in the following manner:
\begin{equation} \label{eq:stb}
	\mathcal{L}_{stb}=|f^{(T)}(\alpha_1, \alpha_2, \beta_1, \beta_2) - C^{(T)}|.
\end{equation}

\textbf{Constraints from the pre-training dataset.} Besides stabilizing the head pose predictions, leveraging information from the \textit{pre-training dataset} is also found beneficial. Its information is learned by minimizing the angular error between the gaze predictions $\mathbf{g}_{pre}$ and the labels $\mathbf{y}_{pre}$.
\begin{equation} \label{eq:sg}
	\mathcal{L}_{pre}(\mathbf{x}_{pre};\Theta)=\langle \mathbf{g}_{pre}, \mathbf{y}_{pre}\rangle.
\end{equation}
Our method is still unsupervised since the actual adaptation data (dual-view data) is unlabeled.

\textbf{Total loss.} The total loss is computed by summing the aforementioned loss functions, as described below:
\begin{equation} \label{eq:total}
	\mathcal{L}=\mathcal{L}_{mut}+\lambda_1\mathcal{L}_{stb}+\lambda_2\mathcal{L}_{pre},
\end{equation}
empirically, we set $\lambda_1=50$ and $\lambda_2=10$.

\subsection{One-to-Two Views Adaptation Procedure}
The unsupervised 1-to-2 views adaptation is outlined in Algorithm~\ref{alg:adaptation}. We start with a pre-trained single-view model $G(\cdot|\Theta^{(1)})$. Labeled images ($\mathcal{D}_{pre}$) from the pre-training dataset and unlabeled images from the dual cameras ($\mathcal{D}$) are input. During UVA, model $G$ is adapted by minimizing \eqnref{eq:total}. The momentum variable $C$ is updated using \eqnref{eq:momentum}.

\textbf{Training details.} We use PyTorch on an NVIDIA 3090 GPU. Pre-training employs the Adam optimizer at a learning rate of $10^{-4}$, while UVA uses $10^{-5}$.

\begin{algorithm}[t]
\caption{Unsupervised 1-to-2 views adaptation.}
\label{alg:adaptation}
\textbf{Input}: {Dual-camera input $\mathcal{D}$, pre-training dataset $\mathcal{D}_{pre}$ and $G$ pre-trained on $\mathcal{D}_{pre}$}\\
\textbf{Output}: {$G(\cdot|\Theta)$}

\begin{algorithmic}[1]
\small	
\STATE Initialize: $C^{(1)} \gets G(\mathbf{x}_1, \mathbf{x}_2|\Theta^{(1)})$ 
\FOR{$ T{\gets}1\; to\; N $}

    \STATE ($\mathbf{x}_{pre}$, $\mathbf{y}_{pre}$),\; $(\mathbf{x}_1, \mathbf{x}_2)$ $\gets$ $\mathcal{D}_{pre}$,\; $\mathcal{D}$
    
    \STATE $p, \mathbf{\hat{y}}$ $\gets$ $G(\mathbf{x}_1, \mathbf{x}_2|\Theta^{(T)})$ with \eqnref{eq:p-y}.
    
    \STATE $\mathcal{L}_{mut}$ $\gets$ $p, \mathbf{\hat{y}}$,\; $G(\mathbf{x}_1, \mathbf{x}_2|\Theta^{(T)})$ with \eqnref{eq:mut}.
    
    \STATE $f^{(T)}$ $\gets$ $G(\mathbf{x}_1, \mathbf{x}_2|\Theta^{(T)})$ with \eqnref{eq:constant}.
    
    \STATE $\mathcal{L}_{stb}$ $\gets$ $f^{(T)}$,\; $C^{(T)}$ with \eqnref{eq:stb}.
    
    \STATE $\mathcal{L}_{pre}$ $\gets$ $\mathbf{y}_{pre}$,\; $G(\mathbf{x}_{pre}|\Theta^{(T)})$ with \eqnref{eq:sg}.
    
    \STATE Train $G(\cdot|\Theta^{(T+1)})$ with \eqnref{eq:total}.
    
    \STATE Update $C^{(T+1)}$ with \eqnref{eq:momentum}

\ENDFOR
\end{algorithmic}
\end{algorithm}


\begin{figure}
	\centering
	\includegraphics[width=0.8\linewidth]{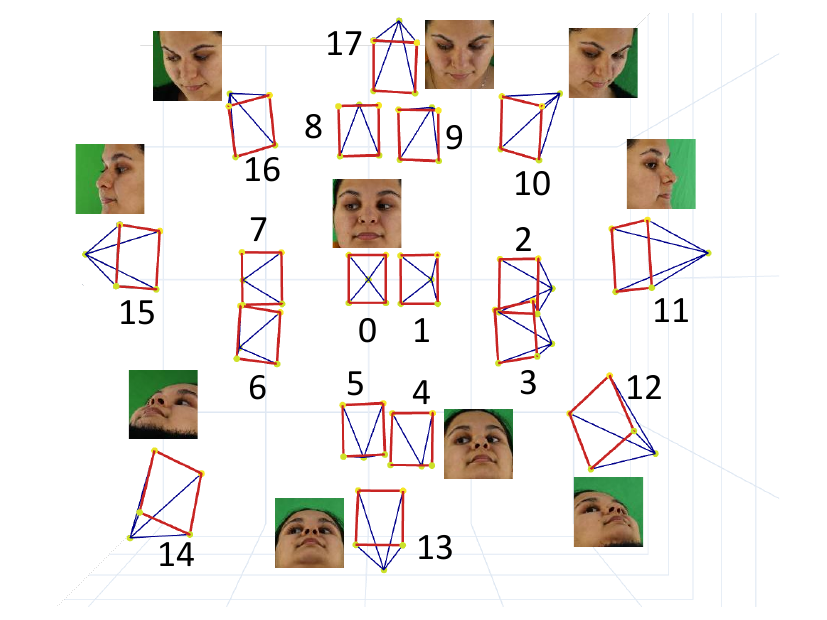}
	\caption{The visualization of the postures of the 18 cameras in ETH-XGaze and their captured image samples.} 
	\label{fig:cameras}
\end{figure}

\section{Data Preparation}\label{sec:preparation}
\textbf{Dual-camera division.} We pre-train using two datasets: ETH-XGaze \cite{G:zhang2020eth} and Gaze360 \cite{G:kellnhofer2019gaze360}. But only the ETH-XGaze dataset is for testing, since only it provides multi-view data. We split its 18 cameras into 9 dual-camera pairs (camera 0-9; 1-10; 2-11; ...). This simple division strategy is chosen given the vast number of potential pair combinations. The positions of the 18 cameras are visualized in \cref{fig:cameras}. 

\textbf{Fine-tuning the ETH-XGaze.} The original ETH-XGaze dataset doesn't fully adhere to the intrinsic consistency of gaze directions. That is, after transforming to the head coordinate system, the gaze directions aren't strictly identical. We find it is due to the inaccurate head pose labels. To rectify this, we refine the head pose labels, starting with the fine-tuning of the head position, denoted as $\mathbf{t}$:
\begin{equation} \label{eq:position}
	\mathbf{t}_{new} = \frac{1}{18}\sum_{i=0}^{17} \mathbf{t}_i,
\end{equation}
\noindent next, we fine-tune the rotation angles, denoted as $\mathbf{h}$, by adding a correction term $\Delta\mathbf{h}$, which is determined as:
\begin{equation} \label{eq:rotation}
	\underset{\Delta\mathbf{h}}{\arg\min}\; \underset{0\leq i\leq 17}{\mathbf{var}}[R^T(\mathbf{h}_i+\Delta\mathbf{h}_i) \times \mathbf{g}_i] + \delta\sum_{i=0}^{17}|\Delta \mathbf{h}_i|,
\end{equation}
\noindent where the $\mathbf{var}$ indicates the variance, and the gaze direction $\mathbf{g}$ is calculated as the vector from head position $\mathbf{t}$ to the gaze target point. We have named the fine-tuned dataset as ETH-MV (Multi-View), which is the foundation for all our experiments. The ETH-MV is available at the project page.

\textbf{Image normalization.} We adopt the commonly-used approach \cite{G:zhang2017mpii} to normalize the data. The normalization aims to ensure consistency in the form of all input images by eliminating head rolls and translating all facial images to the same distance from the camera. To achieve this, the images are wrapped and cropped using an affine transformation matrix $W$, derived from the head pose and camera focal length. 

\begin{table*}[t]
\begin{center}
    \small
    \renewcommand\arraystretch{0.78}
    \setlength{\tabcolsep}{1mm}
    \begin{tabular}{cl|ccc|c||ccc|c}
        \toprule[1.2pt]
        \multirow{2}{*}{Camera pair} & \multirow{2}{*}{Method} & \multicolumn{4}{c}{\underline{Cross-dataset (Gaze360$\rightarrow$ETH)}} & \multicolumn{4}{c}{\underline{In-dataset (ETH$\rightarrow$ETH)}} \\
         & & \mono & \so  & \avg & \hp~(ref) & \mono & \so  & \avg & \hp~(ref) \\
        
        \hline
        \specialrule{0em}{1pt}{1pt}
        
        \multirow{2}{*}{cam 0,9}
        & Baseline        & 13.57 & 11.70 & 13.90 & \textbf{5.44} & 3.55 & 3.37 & 3.47 & 1.40 \\
        & \textbf{\proposed} & \textbf{8.75} \upscore{35.5} & \textbf{6.69} \upscore{42.8} & \textbf{9.45} \upscore{32.0} & 8.13 \downscore{49.6} & \textbf{3.03} \upscore{14.5} & \textbf{2.68} \upscore{20.6} & \textbf{2.89} \upscore{16.7} & \textbf{1.35} \upscore{3.3} \\\cline{1-10} \specialrule{0em}{1pt}{1pt}
        
        \multirow{2}{*}{cam 1,10}
        & Baseline        & 15.06 & 12.54 & 15.03 & \textbf{5.49} & 4.27 & 3.48 & 3.87 & 1.76 \\
        & \textbf{\proposed} & \textbf{10.43} \upscore{30.7} & \textbf{7.11} \upscore{43.3} & \textbf{10.10} \upscore{32.8} & 7.80 \downscore{42.2} & \textbf{4.05} \upscore{5.2} & \textbf{2.87} \upscore{17.5} & \textbf{3.81} \upscore{1.5} & \textbf{1.69} \upscore{4.4}  \\\cline{1-10} \specialrule{0em}{1pt}{1pt}
        
        \multirow{2}{*}{cam 2,11}
        & Baseline        & 18.03 & 16.61 & 19.53 & \textbf{7.68} & 5.66 & 4.42 & 5.25 & 1.60 \\
        & \textbf{\proposed} & \textbf{13.13} \upscore{27.2} & \textbf{10.26} \upscore{38.2} & \textbf{15.39} \upscore{21.2} & 9.76 \downscore{27.1} & \textbf{4.68} \upscore{17.4} & \textbf{3.43} \upscore{22.4} & \textbf{4.20} \upscore{19.9} & \textbf{1.48} \upscore{7.5}\\\cline{1-10} \specialrule{0em}{1pt}{1pt}
        
        \multirow{2}{*}{cam 3,12}
        & Baseline        & 22.85 & 18.34 & 23.20 & \textbf{7.72} & 4.66 & 4.54 & 4.55 & \textbf{1.76} \\
        & \textbf{\proposed} & \textbf{14.93} \upscore{34.7} & \textbf{9.99} \upscore{45.4} & \textbf{15.48} \upscore{33.3} & 8.22 \downscore{6.5} & \textbf{4.30} \upscore{7.7} & \textbf{3.79} \upscore{16.5} & \textbf{4.41} \upscore{3.1} & 1.96 \downscore{11.0} \\\cline{1-10} \specialrule{0em}{1pt}{1pt}

        \multirow{2}{*}{cam 4,13}
        & Baseline        & 24.27 & 21.70 & 24.24 & \textbf{6.06} & 4.66 & 4.01 & 4.10 & 1.40 \\
        & \textbf{\proposed} & \textbf{16.87} \upscore{30.4} & \textbf{13.39} \upscore{38.3} & \textbf{17.74} \upscore{26.8} & 6.75 \downscore{11.4} & \textbf{4.09} \upscore{12.1} & \textbf{3.50} \upscore{12.7} & \textbf{3.63} \upscore{11.4} & \textbf{1.36} \upscore{2.7} \\\cline{1-10} \specialrule{0em}{1pt}{1pt}
        
        \multirow{2}{*}{cam 5,14}
        & Baseline        & 24.03 & 17.55 & 27.19 & \textbf{9.57} & 5.42 & 3.93 & 5.02 & \textbf{1.56} \\
        & \textbf{\proposed} & \textbf{13.76} \upscore{42.7} & \textbf{8.64} \upscore{50.7} & \textbf{20.99} \upscore{22.8} & 10.26 \downscore{7.3} & \textbf{4.62} \upscore{14.9} & \textbf{3.18} \upscore{19.2} & \textbf{4.39} \upscore{12.6} & 1.74 \downscore{11.8} \\\cline{1-10} \specialrule{0em}{1pt}{1pt}
        
        \multirow{2}{*}{cam 6,15}
        & Baseline        & 18.57 & 17.31 & 20.61 & \textbf{7.29} & 5.19 & 3.81 & 4.44 & 1.51 \\
        & \textbf{\proposed} & \textbf{14.00} \upscore{24.6} & \textbf{11.20} \upscore{35.3} & \textbf{16.62} \upscore{19.4} & 8.22 \downscore{12.7} & \textbf{4.78} \upscore{7.9} & \textbf{3.24} \upscore{15.0} & \textbf{4.20} \upscore{5.5} & \textbf{1.48} \upscore{2.4} \\\cline{1-10} \specialrule{0em}{1pt}{1pt}
        
        \multirow{2}{*}{cam 7,16}
        & Baseline        & 17.09 & 14.95 & 17.61 & \textbf{6.77} & 4.99 & 3.93 & 4.37 & \textbf{1.65} \\
        & \textbf{\proposed} & \textbf{12.48} \upscore{27.0} & \textbf{9.98} \upscore{33.2} & \textbf{12.32} \upscore{30.0} & 7.66 \downscore{13.1}& \textbf{4.26} \upscore{14.7} & \textbf{3.30} \upscore{16.0} & \textbf{3.88} \upscore{11.2} & 1.81 \downscore{9.3}  \\\cline{1-10} \specialrule{0em}{1pt}{1pt}
        
        \multirow{2}{*}{cam 8,17}
        & Baseline        & 15.10 & 13.82 & 14.95 & \textbf{5.46} & 4.36 & 3.96 & 4.06 & \textbf{1.63}\\
        & \textbf{\proposed} & \textbf{13.05} \upscore{13.6} & \textbf{12.10} \upscore{12.4} & \textbf{12.40} \upscore{17.0} & 5.93 \downscore{8.8} & \textbf{4.06} \upscore{6.8} & \textbf{3.47} \upscore{12.4} & \textbf{3.83} \upscore{5.6} & 1.69 \downscore{4.0} \\\cline{1-10} \specialrule{0em}{1pt}{1pt}
        
        \multirow{2}{*}{Overall}
        & Baseline        & 18.73 & 16.06 & 19.58 & \textbf{6.83} & 4.75 & 3.94 & 4.35 & \textbf{1.59} \\
        & \textbf{\proposed} & \textbf{13.05} \upscore{30.3} & \textbf{9.93} \upscore{38.2} & \textbf{14.50} \upscore{26.0} & 8.08 \downscore{18.3} & \textbf{4.21} \upscore{11.4} & \textbf{3.27} \upscore{16.9} & \textbf{3.92} \upscore{9.9} & 1.62 \downscore{2.0} \\
        
        \bottomrule[1.2pt]
    \end{tabular}
\end{center}
\caption{Unsupervised 1-to-2 views results for 9 dual-camera pairs. The pairs are from the ETH-MV dataset. In “In-dataset” and “Cross-dataset” settings, the baseline model (ResNet18 \cite{N:he2016deep}) is pre-trained on Gaze360 or ETH-MV, respectively. The adaptation yields a uniquely adapted model for each pair, and “Overall” averages the results across all nine pairs. Note that the head pose error (\hp) is only for reference.}
\label{tab:gaze360-pair}
\end{table*}

\section{Experiments}

\subsection{Metric Definition}
In this paper, we study the 1-to-2 views UVA task, which adapts a single-view gaze estimator for flexible dual cameras. Therefore, traditional monocular angular error (\mono) metric is insufficient. Inspired by the two approaches presented in \secref{sec:intuitive}, we propose two new dual-view metrics, select-front error (\so) and average error (\avg).


It is important to note that the average error can be significantly affected by the head pose error. To account for this, we also measure the head pose error (\hp) using mean-angle-error for reference. 

In summary, we use four metrics to evaluate performance on the dual-view task in this paper, namely:
\begin{itemize}
\setlength{\itemsep}{0pt}
\setlength{\parsep}{0pt}
\setlength{\parskip}{0pt}
\item \mono: the traditional monocular angular error, which collects all the single-view errors from both views and calculates their average.
\item \so: the error under dual-view, calculated by selecting the input image with a smaller head angle.
\item \avg: the error under dual-view, calculated by averaging the predictions from both views.
\item \hp: the mean error across the three angles of head pose ($\alpha,\beta,\gamma$). Note that this is only a reference metric and not a measure of performance.
\end{itemize}

\subsection{One-to-Two Views UVA Results for Flexible Dual Views}
In this section, we use the \proposed~to adapt a single-view estimator for each of the 9 dual-camera pairs from ETH-MV.
The estimator is pre-trained on either the Gaze360 \cite{G:kellnhofer2019gaze360} or ETH-MV dataset, leading to two distinct tasks: Gaze360-to-ETH and ETH-to-ETH. The same baseline model is adapted to each dual-camera pairs with unlabeled data independently, yielding 9 adapted models. Each adapted model is then tested independently.

The results are presented in \cref{tab:gaze360-pair}. Compared to the pre-trained model (Baseline), our method shows significant improvements in both tasks for all dual-camera pairs, as measured by the three gaze accuracy metrics (\mono, \so, and \avg). This indicates that our method is effective for flexibly placed dual camera pairs. There's a slight increase in the head pose error (\hp), primarily because our method does not impose strong constraints on it.

Due to the inclusion of the head pose error (\hp) in the average of gaze predictions, \avg~is larger than the \so~in both tasks. This suggests that using the select-front strategy is optimal when accurate head poses are unobtainable.

\subsection{One-to-Two Views UVA Results under Different Head Angles}
\begin{figure}[t]
\centering
\includegraphics[width=\linewidth]{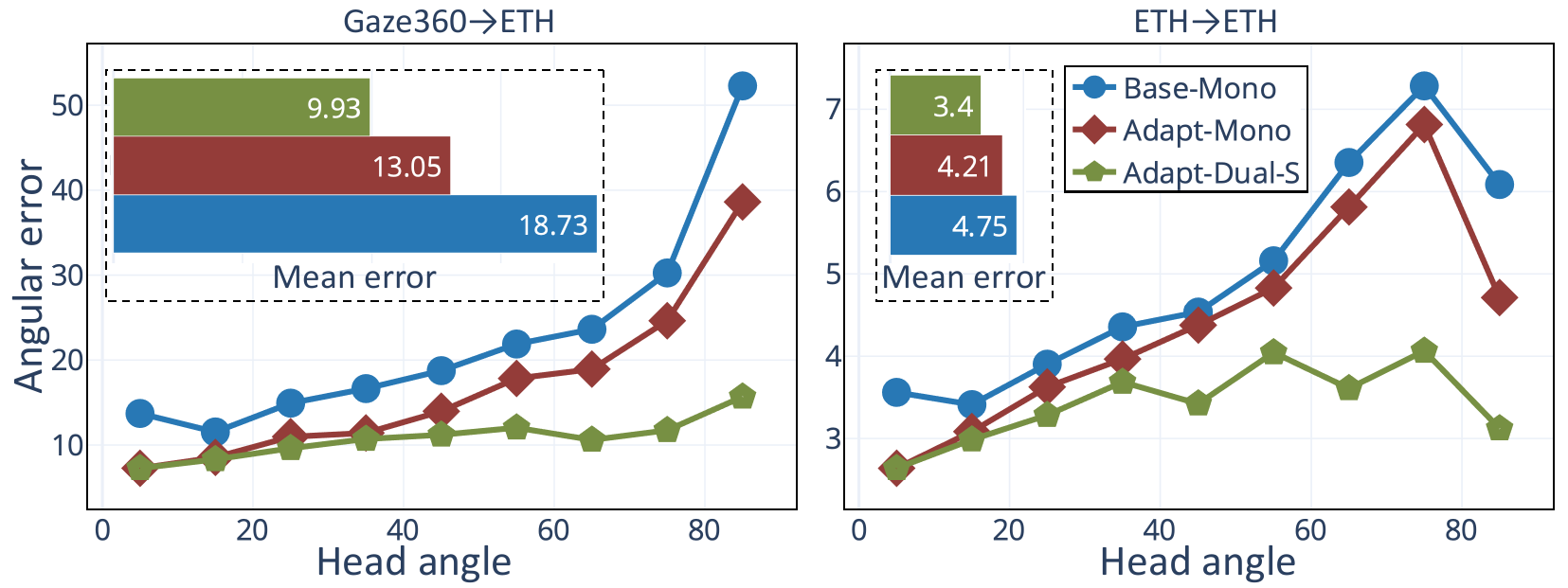}
\caption{The blue and red lines show monocular gaze errors with respect to the head angle on two tasks before and after adaptation, respectively. The green line shows the errors after selecting the predictions with smaller head angles.}
\label{fig:overall}
\end{figure}

To verify the performance of the \proposed~across various head angles, we conduct experiments using the overall error across all dual-camera pairs, illustrating accuracy variations with respect to the head angles.
As depicted in \cref{fig:overall}, the monocular gaze error (Adapt-\mono) of the adapted model increases with increasing head angle. However, there's a notable improvement in accuracy after adaptation.

Remember, our method is grounded in dual-view inference, allowing us to combine predictions from both views using strategies such as the select-front approach. After applying this strategy, the select-front error (\so, green lines in \cref{fig:overall}) is reduced to the level of 0-10$^\circ$ of head angle before adaptation. This indicates that our method can largely eliminate the negative effects caused by large head angles.

\subsection{Ablation Study}
\label{sec:ablation}

An ablation study is conducted to evaluate the effectiveness of each component in \proposed. The components are:
\begin{itemize}
\setlength{\itemsep}{0pt}
\setlength{\parsep}{0pt}
\setlength{\parskip}{0pt}
\item Baseline: a standard CNN-based gaze estimation network using architecture of ResNet18 \cite{N:he2016deep}.
\item mut: the mutual supervision module, which uses the intrinsic consistency of gaze directions under both views to supervise the adaptation.
\item stb: the momentum head pose stabilization module, which uses the fixed rotation transformation between the two camera coordinate systems for stabilization.
\item pre: the gaze constraint that leverages the information from the pre-training dataset.
\end{itemize}

We base our experiments on the Gaze360-to-ETH task for clearer observation of differences. Results are shown in \cref{tab:abla}. Notably, the results for ``CNN+mut" reveal that error amplification occurs if the model uses its own predictions for mutual supervision without any other constraint.

In contrast, with the momentum head pose stabilization module (CNN+mut+stb), the mutual supervision module achieves a significant improvement over the baseline. Additionally, adding the gaze constraint from the pre-training dataset can also stabilize the adaptation process (CNN+mut+pre) and provide additional accuracy improvement (CNN+mut+stb+pre). Ultimately, we select the ``CNN+mut+stb+pre" combination as our final version.

\begin{table}
\small
\setlength{\tabcolsep}{1mm}{
\begin{center}
\renewcommand\arraystretch{0.8}
\begin{tabular}{lcccc}
\toprule[1.2pt]
Method & \mono & \so & \avg & \hp \\
\hline  \specialrule{0em}{1pt}{1pt}
Baseline (w/o mut) & 18.73 & 16.06 & 19.58 & \textbf{6.83} \\
\hline  \specialrule{0em}{1pt}{1pt}
CNN+mut$^\dag$         & 57.73 & 57.09 & 56.98 & 13.67 \\
CNN+mut+stb            & 17.43 & 14.81 & 18.62 & 7.60\\
CNN+mut+pre             & 13.33 & 10.23 & 15.21 & 9.95 \\
CNN+mut+stb+pre         & \textbf{13.05} & \textbf{9.93} & \textbf{14.50} & 8.08\\

\bottomrule[1.2pt]
\end{tabular}
\end{center}
}
\caption{Ablation study results of our dual-view adaptation method on the Gaze360-to-ETH task. $^\dag$ indicates error amplification occurs. 
}
\label{tab:abla}
\end{table}

\begin{figure}[t]
    \centering
    \begin{minipage}{0.48\linewidth}
        \centering
        \includegraphics[width=\linewidth]{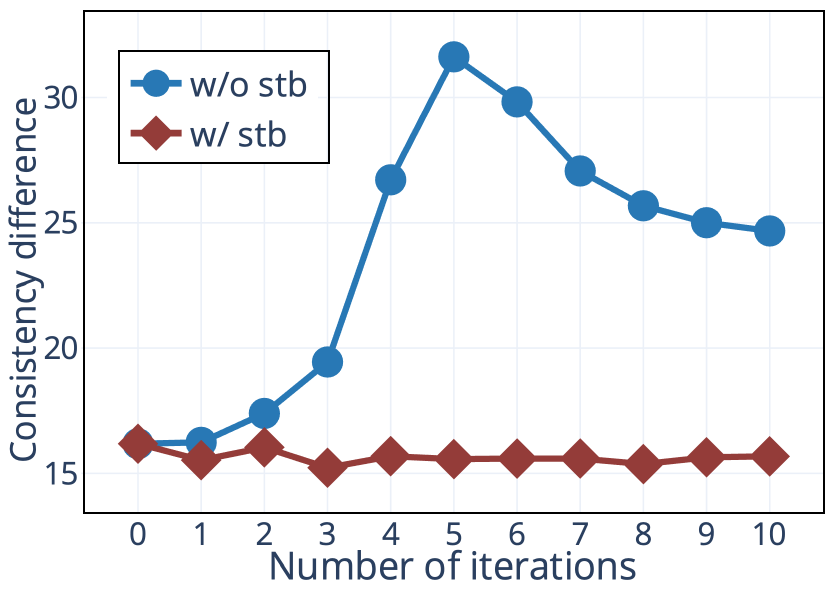}
	\caption{Comparison of the consistency between our method with/without the head pose stabilization module, lower is better.}
        \label{fig:consistency}
    \end{minipage}\hfill
    \begin{minipage}{0.48\linewidth}
        \centering
        \includegraphics[width=\linewidth]{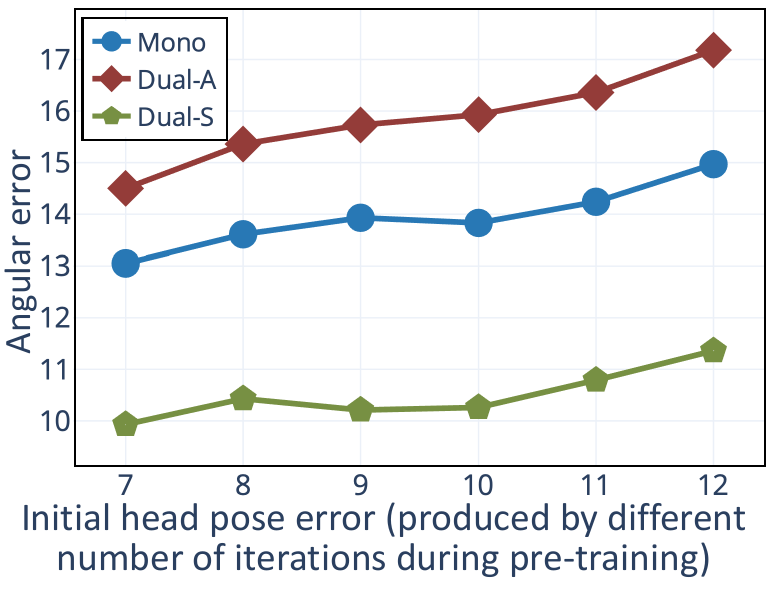}
	\caption{Adaptation results of \proposed~for pre-trained models with different initial head pose accuracy.}
        \label{fig:character}
    \end{minipage}
\end{figure}


\subsection{Effect of the Head Pose Stabilization Module}
The ablation study indicates that the head pose stabilization module can prevent error amplification. Here, we delve deeper into the module's impact during adaptation.

Our experiments compare the gaze prediction consistency between the two views, both with and without the stabilization module. To measure the consistency, first, we transform the two gaze predictions to the same head coordinate system using the head pose predictions. We then define the angular difference between them as the consistency metric, \ie, $\langle R_1^T\mathbf{g}_1, R_2^T\mathbf{g}_2 \rangle$.
The comparison results are shown in \cref{fig:consistency}. The consistency metric between the two views increases sharply without the stabilization module, highlighting a failure in dual-view consistency. In contrast, the stabilization module ensures consistency even with a slight improvement, indicating why this module prevents error amplification.

\subsection{System Characteristics Analysis}

In our experiments, we observe that the initial head pose error of the pre-trained model has a significant impact on the final adaptation result. In this section, we compare the adaptation results of \proposed~for pre-trained models with different initial head pose errors.

We choose several models with different head pose errors from their pre-training phase and then apply the \proposed~for UVA. The adapted models are then tested on the Gaze360-to-ETH task, as shown in \cref{fig:character}.
Clearly, our method enhances accuracy across the board, irrespective of the head pose accuracy variations in the pre-trained model (Baseline: 18.73). Moreover, higher accuracy in head pose estimation directly correlates with higher adaptation results. Hence, in practical, we recommend opting for a pre-trained model with the highest accuracy in head pose estimation.

\subsection{Comparison under Cross-Dataset Settings}
\begin{table}
\small
\centering
\renewcommand\arraystretch{0.85}
\setlength{\tabcolsep}{0.4mm}
\begin{tabular}{lcccccc}  
\toprule[1.2pt]

& \multicolumn{3}{c}{\underline{Gaze360 $\rightarrow$ ETH}} & \multicolumn{3}{c}{\underline{ETH $\rightarrow$ ETH}}  \\ 
& \mono   & \so  & \avg & \mono   & \so  & \avg \\ 
\midrule[1pt]

Baseline	 	& 18.73 & 16.06 & 19.58 & 4.75 & 4.08 & 4.35 \\
\midrule[1pt] 
PnP-GA & 26.36 & 18.60 & 26.23 & 4.63 & 4.10 & 4.31 \\
RUDA & 14.01 & 10.29 & 15.40 & 4.60 & 3.62 & 5.31 \\
DAGEN & 16.26 & 13.48 & 16.58 & 5.14 & 4.48 & 4.75 \\
Gaze360 & 17.10 & 12.44 & 17.53 & 5.78 & 4.60 & 8.07 \\
GazeAdv & 15.00 & 11.12 & 15.93 & 9.09 & 7.89 & 11.00 \\
\textbf{\proposed~(Ours)} & \textbf{13.05} & \textbf{9.93} & \textbf{14.50} & \textbf{4.21} & \textbf{3.40} & \textbf{3.92} \\
\bottomrule[1.2pt]
\end{tabular}
\caption{
Comparison with other approaches that have state-of-the-art performance under cross-dataset settings. 
}
\label{tab:sota}
\end{table}

In this section, we compare our method with leading techniques that demonstrate state-of-the-art results in cross-dataset scenarios. Given the practical significance of cross-dataset capability, this experiment is designed to evaluate the capability of our \proposed~in comparison to leading single-view methods. Specifically, the methods under comparison include PnP-GA \cite{D:liu2021generalizing}, RUDA \cite{D:bao2022generalizing}, DAGEN \cite{D:guo2020dagen}, Gaze360 \cite{G:kellnhofer2019gaze360}, and GazeAdv \cite{D:wang2019gazeadv}.



The comparison is shown in \cref{tab:sota}. For the Gaze360-to-ETH task, our method has the best performance. Conversely, the PnP-GA method experiences error amplification when applied to the task, and the performance of other methods pales in comparison to our \proposed.

In addition, we present results for the ETH-to-ETH task, even though this is an in-dataset task. It is shown that only our method has a consistent improvement across all three metrics. Evidently, \proposed~continues to surpass the performance of all the existing methods.



\section{Conclusion}
In this paper, we introduce the \proposed, a groundbreaking 1-view-to-2-views adaptation framework for gaze estimation. This method adapts a traditional single-view gaze estimator for flexible dual views. Central to \proposed~is a mutual supervision strategy, which takes advantage of the intrinsic consistency of the gaze directions. Notably, our method demonstrates superior performance when adapting a single-view estimator for dual-view scenarios.


\bibliography{aaai24.bib}

\end{document}